\documentclass[11pt]{article}

\usepackage[utf8]{inputenc}
\usepackage[T1]{fontenc}
\usepackage{times}
\usepackage[margin=1in]{geometry}
\usepackage{microtype}
\usepackage{amsmath,amssymb}
\usepackage{graphicx}
\usepackage{booktabs}
\usepackage{array}
\usepackage{enumitem}
\usepackage{subcaption}
\usepackage{xcolor}
\usepackage{pgfplots}
\pgfplotsset{compat=1.16}
\usepackage[numbers,sort&compress]{natbib}
\usepackage[colorlinks=true,linkcolor=blue!50!black,citecolor=blue!50!black,urlcolor=blue!50!black]{hyperref}
\usepackage{url}

\newcommand{\NAS}{\mathrm{NAS}}
\newcommand{\KIS}{\mathrm{KIS}}
\newcommand{\IK}{\textsc{ik}}
\newcommand{\OK}{\textsc{ok}}

\title{\textbf{Confident Fools or Careful Searchers?\\A Behavioural Study of When LLMs Search\\Beyond Their Static Internal Memory}}

\author{
  Nipun Pathirage \\
  Department of Computer Science\\
  Rensselaer Polytechnic Institute, Troy, NY, USA\\
  \texttt{pathin@rpi.edu}
  \and
  Kwan Srijomkwan \\
  Department of Cognitive Science\\
  Rensselaer Polytechnic Institute, Troy, NY, USA\\
  \texttt{srijok@rpi.edu}
}
\date{}

\begin{document}
\maketitle

\begin{abstract}
Agentic LLM assistants must repeatedly decide whether to answer from memory or to call a tool such as web search. Unnecessary searches cost latency and money; missing ones yield confident but unsupported answers. Prior work mostly measures verbalised confidence or trains new adaptive-retrieval methods, and rarely audits the search decisions that deployed products actually make. We introduce a benchmark of 400 multiple-choice questions that combines widely known inside-knowledge (\IK) questions, which should be answered directly, with outside-knowledge (\OK) questions about facts from news published after the models' training cut-offs (to our knowledge), which should trigger search or, without tools, abstention. Each item offers an abstain option, and two action-level scores, NAS and KIS, reward searching on \OK{} questions and restraint on \IK{} questions. Across five commercial LLMs with and without their vendors' web search, GPT-5 mini and Grok 4 Fast search on over 90\% of \OK{} but only 1\% of \IK{} questions, whereas Mistral Small searches on just 20\% of \OK{} questions, often abstaining instead of using its tool; one product's search indicator proved unusable. Searching is partly triggered by explicit temporal cues: on uncued \OK{} questions, the NAS of the two selective models falls from about 0.72 to 0.57--0.58. The scores thus measure the search triggering of specific products, not metacognition in a strict sense.
\end{abstract}

\section{Introduction}

Consider an assistant asked two questions in succession: which international agreement regulates the trade in conventional arms, and which precious metal fell by Rs~3{,}900 per 10 grams in New Delhi on 18 November 2025. The first is textbook knowledge held in any capable LLM's parameters; the second comes from a single day's market report that no model saw during training. An agentic system with web search \citep{yao2023react,xi2025rise} should answer the first directly and search before answering the second. Both kinds of error are costly. Searching on everything is slow and expensive and adds noisy retrieval for no benefit; searching too rarely produces fluent answers from missing knowledge, exactly where hallucinations arise \citep{ji2023survey,huang2025survey}. The decision to search is a small act of self-assessment that a tool-using assistant makes on every turn.

This decision has a metacognitive flavour. Psychology calls the ability to monitor and regulate one's own cognition \emph{metacognition} \citep{flavell1979metacognition}, and the Dunning--Kruger effect, in which people with low competence overestimate it, is the best-known illustration of failed self-assessment \citep{kruger1999unskilled}. We use this vocabulary as a lens, not as a hypothesis. Human studies compare self-reported confidence with measured performance across a population, and we do not test that pattern in LLMs. Instead, we treat a model's \emph{observable action} (answering, searching or abstaining) as an expression of its implicit judgement about whether a question lies within its knowledge, and ask whether that action fits the question.

Existing work approaches the problem from two directions. One line asks whether LLMs know what they do not know, through the calibration of their answer probabilities \citep{kadavath2022mostly}, their recognition of unanswerable questions \citep{yin2023selfaware}, the quality of prompted or fine-tuned uncertainty estimates \citep{kapoor2024taught}, and abstention \citep{wen2025abstention}. These studies measure confidence in a single answer, not what a system does when a tool is available. A second line, motivated by evidence that retrieval helps most on long-tail and fast-changing knowledge \citep{mallen2023trust,vu2024freshllms}, develops methods that retrieve or call tools only when needed \citep{asai2024selfrag,li2025meco} or that train agents to avoid tool overuse \citep{qian2025smart}. These works evaluate new methods in their own settings. What is missing is a simple way to audit the search decisions of the systems people actually use: off-the-shelf commercial LLMs calling their vendors' own search tools, tested on both sides of their knowledge boundary.

We build such an audit around two kinds of questions. Outside-knowledge (\OK) questions ask about a specific fact from news published on 17--18 November 2025, after the evaluated models' training cut-offs, to our knowledge; with a search tool the right action is to search, and without one it is to abstain. Inside-knowledge (\IK) questions are generic questions on related themes with widely known answers; the right action is to answer directly. Every item has a fifth ``not enough information'' option, so abstention is always available and objectively scorable. We run five commercial models with and without their vendors' built-in web search and summarise their actions with two scores in $[-1,1]$. The Net Awareness Score (NAS), on \OK{} questions, counts the items on which a model acts on its knowledge gap (for example, searching and then answering correctly) minus those on which it ignores the gap (for example, not searching at all). The Knowledge Integrity Score (KIS), on \IK{} questions, counts correct answers given without search minus needless searches and abstentions. A model with high NAS and high KIS searches when, and only when, it needs to.

The scores separate the products clearly. GPT-5 mini and Grok 4 Fast search on 90--92\% of \OK{} questions but on only 2 of 200 \IK{} questions, and reach 0.83 accuracy on \OK{} questions with search. Mistral Small fails in the opposite direction: when it searches it is nearly always right (38 of 39), but it does not search on 152 of 191 \OK{} questions and abstains on 90 of these, giving a NAS of $-0.51$ alongside a KIS of 0.80; it seems to notice the gap without acting on it. Even without tools, all models abstain on 43--77\% of \OK{} questions but on at most 0.5\% of \IK{} questions. One qualification matters: about 58\% of \OK{} stems contain an explicit temporal cue such as a year or ``recent'', against 1 of 200 \IK{} stems, and on the 83 uncued \OK{} questions the NAS of GPT-5 mini and Grok 4 Fast falls to 0.57--0.58, from 0.83--0.85 on cued ones. Part of what looks like self-knowledge is sensitivity to wording, so we interpret the scores as measures of \emph{search triggering}. The study is also preliminary: it compares products with different search stacks, uses one run per item, and cannot reliably interpret the with-search behaviour of two of the five (Gemini 2.5 Flash and Claude 3.5 Haiku).

Our contributions are:
\begin{itemize}[leftmargin=*,itemsep=1pt,topsep=2pt]
  \item A 400-item multiple-choice benchmark combining inside- and outside-knowledge questions derived from news published after the models' cut-offs, each with an explicit abstention option.
  \item Two action-level scores, NAS and KIS, that rate search, answer and abstention decisions on each side of a model's knowledge boundary, with and without a search tool.
  \item An evaluation of five commercial LLMs with and without their vendors' web search, with outcome breakdowns showing how each score arises.
  \item An analysis of what the scores measure, showing that search triggering is partly cue-driven and that KIS, unlike with-search NAS, depends strongly on an LLM judge of reasoning alignment.
\end{itemize}
After reviewing related work, we describe the benchmark (Section~\ref{sec:dataset}), scores (Section~\ref{sec:metrics}) and setup (Section~\ref{sec:setup}), then present results (Section~\ref{sec:results}) and limitations (Section~\ref{sec:limitations}). Data, code and parsed outputs are available at \url{https://github.com/nipdep/DKE-on-LLM}.

\section{Related Work}

\paragraph{Knowing what one does not know.}
Several works ask whether LLMs can recognise the limits of their knowledge. \citet{kadavath2022mostly} show that larger models are reasonably calibrated on multiple-choice questions and can estimate the probability that their own answer is correct (their ``P(IK)'' is unrelated to our \IK{} label). \citet{yin2023selfaware} introduce SelfAware, a set of answerable and unanswerable questions, and find that models fall short of humans in identifying the unanswerable ones. \citet{kapoor2024taught} argue that prompting alone is insufficient and that small amounts of fine-tuning give better uncertainty estimates, and \citet{wen2025abstention} survey abstention as a response to knowledge gaps. \citet{ma2025intrinsic} argue that LLMs have intrinsic metacognition that needs a good probing ``lens.'' These studies mostly measure verbalised or internal confidence on a single answer. We instead measure the \emph{action} an agentic system takes when a tool is available, which is what ultimately determines cost and reliability in deployment.

\paragraph{Adaptive retrieval and tool use.}
Retrieval is most useful for long-tail or fast-changing knowledge \citep{mallen2023trust,vu2024freshllms}, and prior methods train or prompt models to retrieve only when needed \citep{mallen2023trust,asai2024selfrag,jeong2024adaptive,li2025meco}. \citet{qian2025smart} explicitly frame tool overuse as a lack of self-awareness and train agents to reduce it, and MetaTool \citep{huang2024metatool} benchmarks whether LLMs decide to use a tool at all. These works propose \emph{methods} and evaluate on their own settings; we provide a small behavioural benchmark for auditing \emph{off-the-shelf commercial models} through the search tools their vendors ship, without any additional orchestration.

\paragraph{Hallucination and news-based QA benchmarks.}
Multiple-choice formats make evaluation objective \citep{zhang2024mcq}; TruthfulQA \citep{lin2022truthfulqa} uses adversarial distractors, and RealTime QA \citep{kasai2023realtime} asks weekly news questions. Our benchmark builds on the same ideas but contrasts article-specific \OK{} items with generic \IK{} items drawn from the same pool of articles, so that the same model is tested on both sides of its knowledge boundary. MCQ results are also known to be sensitive to option position and selection bias \citep{zheng2024selection}; we did not permute options. For constructing the items we adapt LLM-based MCQ generation pipelines with multi-stage prompting, joint distractor generation and LLM-based validation \citep{maity2024novel,biancini2024multiple,rodriguez2025joint,mucciaccia2025automatic}.

\paragraph{Metacognition measurement in psychology.}
Cognitive science separates first-order performance from ``Type 2'' monitoring performance and quantifies the latter with signal-detection measures such as meta-$d'$ \citep{maniscalco2014signal,fleming2014measure}; \citet{johnson2022metacognition} motivates metacognition for AI safety. Such measures require trial-level confidence ratings. Because the commercial APIs we evaluate expose actions (search calls, abstentions) rather than calibrated confidence, NAS and KIS are behavioural proxies and are not meta-$d'$ measures.

\section{Benchmark}
\label{sec:dataset}

\subsection{Overview and source content}

The benchmark has 400 items: four domains (technology, health, business, entertainment) $\times$ two knowledge types (\IK, \OK) $\times$ 50 questions. \OK{} questions concern a fact specific to a news article; \IK{} questions are generic questions on the article's theme that do not depend on it.

Source articles were retrieved from the GNews API (\url{https://gnews.io}, free tier, truncated full text) and all carry publication dates of 17 or 18 November 2025. After deduplication, 104--156 unique articles remained per domain. This is months after the training cut-offs of the evaluated models, to our knowledge, which makes it unlikely that article-specific facts were in their training data; we cannot rule out that some \OK{} items concern events that were foreshadowed in earlier data (Section~\ref{sec:limitations}).

From each article we extract one \emph{factoid}: a one-sentence, decontextualised factual statement with explicit referring expressions, a time reference if needed, and brief explanations of unfamiliar entities. The \IK{} and \OK{} generators were run as two separate passes over the article pool, so the two sets are only loosely linked: they share 103 source articles, 30 \IK{} and 35 \OK{} items reuse an article already used by another item of the same type, and five question texts appear twice (mostly generic \IK{} questions). Using a single factoid for each \OK{} question gives tight control over the information needed to answer it.

\begin{figure}[t]
  \centering
  \includegraphics[width=0.62\linewidth]{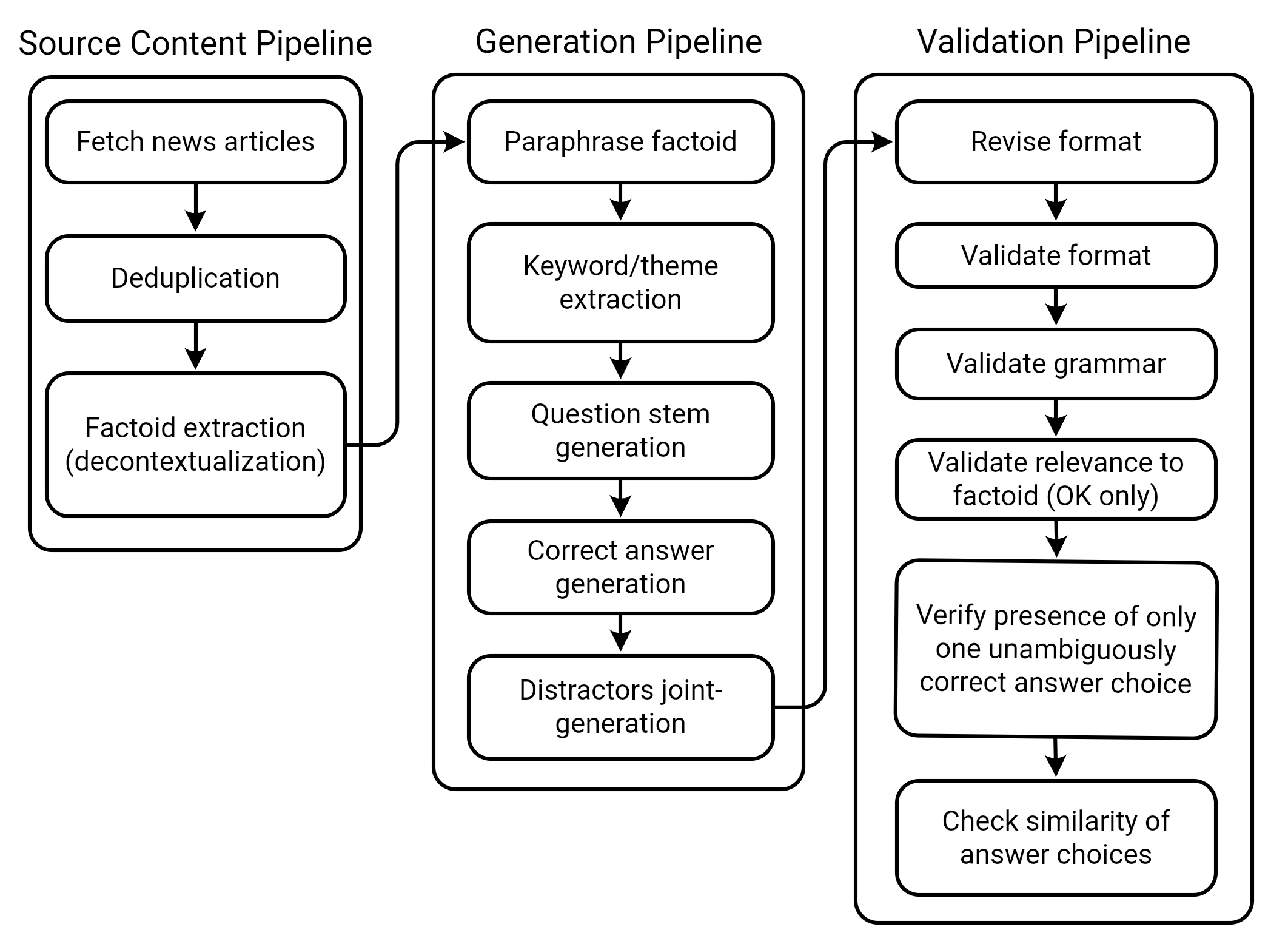}
  \caption{Dataset generation and validation pipeline.}
  \label{fig:pipeline}
\end{figure}

\subsection{MCQ construction}

Questions are generated with GPT-4o-mini (the \emph{generation model}), a model not used for answering, in four stages (Figure~\ref{fig:pipeline}) adapted from \citet{maity2024novel}, with prompts adapted from \citet{biancini2024multiple} (Appendix~\ref{sec:dataset_generation_prompt}): (1) paraphrase the factoid, to reduce lexical overlap with the source; (2) extract a keyword (\OK) or a one-to-two-word theme (\IK); (3) generate the question stem; (4) generate three distractors jointly \citep{rodriguez2025joint}.
\begin{itemize}[leftmargin=*,itemsep=1pt,topsep=2pt]
  \item \textbf{\OK{} items.} The keyword, which is a span of the paraphrased factoid, is the correct answer, and the stem is generated from the factoid and the keyword. Thus the correct answer can be found only in the article's information.
  \item \textbf{\IK{} items.} The generator sees only the theme, not the factoid, and produces a time-insensitive question, a short free-form answer and a one-line rationale. The needed knowledge therefore cannot come from the article. Distractors are asked to match the length of the correct answer, since length mismatches were common.
\end{itemize}
Every item stores a ground-truth rationale used later for reasoning-alignment checks. Examples are in Appendix~\ref{sec:question_examples}.

\subsection{Validation}

All items pass a model-based validation sequence (gpt-5-mini, chosen for cost since each stage is narrow), based on \citet{mucciaccia2025automatic}: (i) format revision of the stem and correct answer to remove stylistic cues that would reveal the answer; (ii) format and grammar checks; (iii) a uniqueness check in which each option is scored 0--10 for correctness independently, and an item is kept only if the keyed option is the top-scoring one with a margin of at least 2 points over the runner-up; (iv) for \OK{} items only, a relevance-to-factoid score $\ge 8/10$; and (v) a cosine-similarity analysis between each distractor and the correct answer (OpenAI \texttt{text-embedding-3-small}). Step (v) is descriptive only: it describes the spread of difficulty and does not remove items. Table~\ref{tab:dataset_stats} summarises the final dataset. The relevance filter is nearly saturated (190 of 200 \OK{} items score 10), so it screens out little; we therefore do not treat it as evidence of high item quality beyond basic relevance.

\begin{table}[t]
\centering
\small
\begin{tabular}{ll}
\toprule
Items & 400 (200 \IK, 200 \OK; 50 per domain and type) \\
Source articles & 17--18 Nov 2025; 104--156 per domain after deduplication \\
Position of keyed option (1/2/3/4) & 102 / 96 / 97 / 105 \\
Relevance score, \OK{} items & mean 9.95; 190 items = 10, 9 items = 9, 1 item = 8 \\
Distractor--answer cosine similarity & median 0.41 (IQR 0.31--0.53); \IK{} 0.44, \OK{} 0.38 \\
Source URLs & 170 (\IK), 165 (\OK); 103 shared; 5 duplicated question texts \\
Explicit temporal cue in stem$^*$ & \OK{} 117/200 (58\%); \IK{} 1/200 \\
\bottomrule
\end{tabular}
\caption{Dataset summary. Score distributions are shown in Appendix Figure~\ref{fig:combined}. $^*$Post-hoc regular-expression match for a year 2024--2029, ``November'', ``recent'', ``latest'', ``current'', ``this week'' or ``today''.}
\label{tab:dataset_stats}
\end{table}

\section{Metrics}
\label{sec:metrics}

We evaluate each model under two conditions: \textbf{with search} (the vendor's built-in web search tool, which the model decides whether to call) and \textbf{without search}. Each response is one of: an answer (options 1--4) or an abstention (option 5), and, in the with-search condition, a binary indicator $s$ of whether search was invoked. Let $y$ indicate that the answer is correct (abstention is incorrect), and $r$ that an LLM judge rates the model's stated reason as aligned with the ground-truth rationale (Section~\ref{sec:setup}); $r$ is only used when $y=1$. Failed runs are excluded and the denominators are the numbers of valid items.

\subsection{Net Awareness Score (NAS), computed on \OK{} items}

For each \OK{} item the response is labelled \emph{aware}, \emph{unaware}, or \emph{neutral}, and
\begin{equation}
  \NAS = \frac{\#\text{aware} - \#\text{unaware}}{\#\text{\OK{} items}}\ \in[-1,1].
\end{equation}
Table~\ref{tab:rules_nas} gives the rules. With search, \OK{} questions \emph{should} trigger search: searching and answering correctly is aware; not searching is unaware even when the model abstains (an agent with a tool is expected to act on the gap), unless it answered correctly with aligned reasoning, which we treat as neutral. Searching but failing to obtain a correct answer is neutral, since the retrieval step may have failed for reasons unrelated to the decision to search. Without search, abstaining is the aware action, a wrong answer is unaware, and a correct answer with aligned reasoning counts as aware, since a correct inference without access to the article is possible and should not be penalised.

\begin{table}[t]
\centering
\small
\begin{tabular}{llp{6.2cm}}
\toprule
Condition & Label & Response pattern \\
\midrule
with search & aware & searched $\wedge$ correct \\
with search & unaware & not searched $\wedge$ (incorrect/abstained $\vee$ correct with misaligned reasoning) \\
with search & neutral & searched $\wedge$ incorrect/abstained; not searched $\wedge$ correct $\wedge$ aligned \\
\midrule
without search & aware & abstained $\vee$ (correct $\wedge$ aligned) \\
without search & unaware & answered $\wedge$ incorrect \\
without search & neutral & answered $\wedge$ correct $\wedge$ misaligned \\
\bottomrule
\end{tabular}
\caption{Per-item labels for NAS on \OK{} questions.}
\label{tab:rules_nas}
\end{table}

\subsection{Knowledge Integrity Score (KIS), computed on \IK{} items}

KIS asks whether the model is confident and correct when it \emph{should} know the answer:
\begin{equation}
  \KIS = \frac{\#\text{confident-correct} - \#\text{wasteful}}{\#\text{\IK{} items}}\ \in[-1,1].
\end{equation}
With search, an item is confident-correct if the answer is correct, aligned, and no search was made; it is wasteful if the model searched or abstained. Without search, an item is confident-correct if correct and aligned, and wasteful if incorrect, abstained, or correct with misaligned reasoning. All other items are neutral. KIS therefore measures \emph{restraint} in the with-search condition but mixes restraint with accuracy and reasoning quality in the no-search condition. This asymmetry is a limitation of the current definition, so we treat the with-search condition (the actual agentic decision) as the primary analysis.

\subsection{Reading the scores}

NAS and KIS capture two complementary failure modes: not acting on a knowledge gap (low NAS), and acting when no action is needed or lacking confidence in known facts (low KIS). Placing a model in the (NAS, KIS) plane gives four informal regimes: \emph{ideal} (high NAS, high KIS), \emph{over-searcher} (high NAS, low KIS), \emph{under-searcher / overconfident} (low NAS, high KIS), and \emph{confused} (both low). The zero thresholds are conventions, not empirical boundaries. We compute 95\% confidence intervals by bootstrapping questions (3{,}000 resamples); because each model is run once on each question, these intervals reflect question sampling only and not run-to-run variability of the models.

\section{Experimental Setup}
\label{sec:setup}

\paragraph{Models.}
We evaluate five commercial models through their APIs (Table~\ref{tab:llms}). Each is run with its vendor's native web-search tool, which the model itself decides whether to invoke, and without any tool. We use the vendor tools unmodified to reflect what users and developers obtain by default; consequently, the search stacks differ across vendors and the comparison is between \emph{products}, not between models in isolation. Mistral Small is open-weight but was accessed through the vendor's API.

\begin{table}[t]
\centering
\footnotesize
\begin{tabular}{>{\raggedright\arraybackslash}p{4.9cm}>{\raggedright\arraybackslash}p{3.7cm}>{\raggedright\arraybackslash}p{3.6cm}c}
\toprule
Model (API id) & Search tool & Search indicator & Indicator \\
\midrule
GPT-5 mini (\texttt{gpt-5-mini-2025-08-07}) & OpenAI web search & web-search call in output & plausible \\
Grok 4 Fast (\texttt{grok-4-fast-reasoning}) & xAI Live Search, \texttt{auto} & citations returned & plausible \\
Claude 3.5 Haiku (\texttt{claude-3-5-haiku-latest}) & Anthropic web search (max 2 uses) & more than one content block in output & unverified \\
Mistral Small (\texttt{mistral-small-2506}) & Mistral Agents web search & extra output entries & plausible \\
Gemini 2.5 Flash (\texttt{gemini-2.5-flash}) & Grounding with Google Search & any response candidate & \textbf{defective} \\
\bottomrule
\end{tabular}
\caption{Evaluated models, their search tools, and how a search is detected. ``Plausible'' means judged plausible by code inspection only; no indicator was validated against raw API responses, which were not stored. The Gemini indicator is defective (see Section~\ref{sec:results}).}
\label{tab:llms}
\end{table}

\paragraph{Prompting.}
The harness (Appendix~\ref{sec:mcq_prompt}) passes every model the same user prompt (the question and five options, the fifth being ``Not enough information to answer'') and the same system prompt (``Answer following multiple choice question based on your knowledge. Provide the answer choice (1, 2, 3, 4 or 5) along with a very short reason for your choice.''), with structured JSON output (\texttt{choice}, \texttt{reason}). In the with-search condition the system prompt is extended with the sentence ``Only use the web search tool if the user asks for current, real-time information or for obscure facts you do not know.'' This instruction is part of the elicitation, and results may differ under other instructions. \emph{However, the prompts actually delivered differ across vendors}, which we found only on auditing the code: for GPT-5 mini the structured-output call (Responses API) receives only the user prompt plus a JSON-formatting sentence, so \emph{GPT-5 mini saw neither the shared system prompt nor the search-discouraging sentence in either condition}; for Grok 4 Fast the with-search system prompt additionally carries a stray, irrelevant sentence about extracting ``invoice data into JSON''. The other models receive the prompts as described. Output length was capped at 100 tokens for the Claude, Grok and Mistral calls, which may limit reasoning text.

\paragraph{Answer pipeline.}
Without search, every model answers in a single structured call. With search, Claude 3.5 Haiku, Gemini 2.5 Flash and Mistral Small answer in two stages: a first call with the search tool produces a free-text answer, and a second call without tools converts that text into the JSON choice and reason (for Claude, the second call uses the same model and is also given the user prompt; for Gemini and Mistral the extractor sees only the first-stage text, not the question or options). The stored reason is therefore that of the second call, and the with-search and no-search conditions differ in pipeline as well as in tool access for these three models. If the first call is truncated or its text is not the final answer, the extraction stage can output an abstention; we cannot rule this out for Claude, whose first-stage text is selected by position from the response content blocks. GPT-5 mini and Grok 4 Fast use a single call in both conditions.

\paragraph{Runs and failures.}
Each question was run once per model and condition (4{,}000 runs). An initial attempt without retries produced roughly 42\% unparseable outputs (mostly format errors; a figure recorded at the time, whose log was not retained), so the runner retries rate-limit errors after a delay and retries errors that its validator classifies as format or parse errors (up to two more times, with a stricter JSON-only instruction appended to the prompt), keeping the first parseable output. Other errors, including some JSON-decoding failures that the validator classifies as unknown, are not retried; all 16 non-rate-limit residual failures below are of this kind. After this, 17 of 4{,}000 runs (0.4\%) still failed: 6 for Claude 3.5 Haiku and 10 for Mistral Small with search, and 1 for Mistral Small without search. These runs are excluded.

\paragraph{Reasoning alignment.}
For correct answers, we use \texttt{gpt-oss-120b}, a model not among those evaluated, to judge whether the model's stated reason matches the ground-truth rationale using a proposition-level rubric (Appendix~\ref{sec:reasoning_prompt}); a score of 70 or more out of 100 is ``aligned.'' This criterion is meant to distinguish a correct inference from a lucky guess. We did not validate the judge against human labels (Section~\ref{sec:limitations}), so Section~\ref{sec:results} reports results with and without it.

\section{Results}
\label{sec:results}

\subsection{Baseline accuracy}

Table~\ref{tab:acc} reports accuracy under the settings that favour each question type. Accuracy on \IK{} questions without search is 98.5--99.5\% for all models, so \IK{} items are essentially at ceiling: they are answerable from parametric knowledge, and any search on them is unnecessary. On \OK{} questions, search raises accuracy substantially for GPT-5 mini and Grok 4 Fast (to 0.83) and, to a lesser degree, for Gemini 2.5 Flash (0.59), while Mistral Small and Claude 3.5 Haiku gain little or nothing relative to no search, for reasons detailed below.

Without search, \OK{} accuracy over all items is low (0.19--0.39) because models frequently abstain. However, accuracy \emph{among answered} \OK{} items is high (0.69--0.79, Table~\ref{tab:acc}), far above the 0.25 chance level. This shows that models that choose to answer often succeed, and also that part of the \OK{} set can be solved by priors or elimination; we return to this in Section~\ref{sec:limitations}.

\begin{table}[t]
\centering
\small
\begin{tabular}{lccccc}
\toprule
 & \multicolumn{1}{c}{\IK, no search} & \multicolumn{2}{c}{\OK, with search} & \multicolumn{2}{c}{\OK, no search} \\
\cmidrule(lr){2-2}\cmidrule(lr){3-4}\cmidrule(lr){5-6}
Model & acc. & acc. & abstain & acc. & acc.\ if answered \\
\midrule
GPT-5 mini        & 0.990 & 0.83 & 0.14 & 0.37 & 0.72 ($n$=101) \\
Grok 4 Fast       & 0.995 & 0.83 & 0.16 & 0.24 & 0.78 ($n$=60) \\
Gemini 2.5 Flash$^\dagger$ & 0.995 & 0.59 & 0.33 & 0.32 & 0.77 ($n$=82) \\
Mistral Small     & 0.990 & 0.42 & 0.47 & 0.39 & 0.69 ($n$=113) \\
Claude 3.5 Haiku  & 0.985 & 0.26 & 0.70 & 0.19 & 0.79 ($n$=47) \\
\bottomrule
\end{tabular}
\caption{Accuracy (abstentions count as incorrect). $n$ is the number of answered \OK{} items without search. 95\% bootstrap intervals are about $\pm$0.01--0.07 for the first column and $\pm$0.05--0.07 for the others (with 191--200 items per cell). $^\dagger$Search indicator unreliable.}
\label{tab:acc}
\end{table}

\subsection{What does each model do on \OK{} questions with search?}

Table~\ref{tab:breakdown} decomposes the responses to \OK{} questions with search into six outcomes, and also gives the search rate on \IK{} questions. This is the most informative view of the data, because it shows how each NAS value arises.

\begin{table}[t]
\centering
\small
\setlength{\tabcolsep}{4pt}
\begin{tabular}{lcccccccc}
\toprule
 & \IK & \multicolumn{3}{c}{\OK: searched (\%)} & \multicolumn{3}{c}{\OK: not searched (\%)} & \\
\cmidrule(lr){3-5}\cmidrule(lr){6-8}
Model & search \% & correct & abst. & wrong & correct & abst. & wrong & $n$ \\
\midrule
GPT-5 mini & 1.0 & 77.0 & 12.0 & 2.5 & 5.5 & 2.0 & 1.0 & 200 \\
Grok 4 Fast & 1.0 & 77.5 & 12.0 & 0.5 & 5.5 & 3.5 & 1.0 & 200 \\
Mistral Small & 0.0 & 19.9 & 0.0 & 0.5 & 22.0 & 47.1 & 10.5 & 191 \\
Claude 3.5 Haiku & 11.0 & 19.1 & 40.7 & 4.1 & 6.7 & 29.4 & 0.0 & 194 \\
Gemini 2.5 Flash$^\dagger$ & 100$^\dagger$ & 59.0 & 33.0 & 8.0 & -- & -- & -- & 200 \\
\bottomrule
\end{tabular}
\caption{Outcomes on \OK{} questions with search (percentage of valid items), and the share of \IK{} questions on which the model searched. $^\dagger$Gemini's search indicator is constant (see the Gemini paragraph below).}
\label{tab:breakdown}
\end{table}

\paragraph{GPT-5 mini and Grok 4 Fast search selectively.}
Both searched on 90--92\% of \OK{} questions and on only 1\% (2 of 200) of \IK{} questions: high recall and high precision of search triggering. (Haiku and Mistral also search more on \OK{} than on \IK{} questions, but with much lower recall.) When GPT-5 mini and Grok searched, they were right on 84--86\% of \OK{} items; about 12\% of \OK{} items end in an abstention even after searching. Inspecting the item structure suggests that some of these are questions that cannot be answered even with search (e.g., those referring to ``the article mentioned'' without naming it) or whose keyed answer is wrong or time-dependent (Section~\ref{sec:limitations}), so the ceiling on \OK{} accuracy is below 1. Recall that GPT-5 mini was not given the search-discouraging instruction (Section~\ref{sec:setup}).

\paragraph{Mistral Small often abstains but rarely searches.}
Mistral searched on only 20\% of \OK{} questions and on none of the \IK{} questions. When it searched, it was nearly perfect (38 of 39 correct). On the 152 \OK{} questions where it did not search, it abstained on 90 (59\%), answered 42 correctly (28\%) and answered 20 incorrectly (13\%). Its NAS of $-0.51$ is thus driven largely by a \emph{failure to act}: it often selects the ``not enough information'' option yet does not invoke the tool that could close the gap. Abstention is only a weak proxy for having noticed the gap, since it could also reflect a general tendency to abstain, but the pattern is consistent with noticing without acting. Because our NAS counts abstention without search as unaware in the with-search condition, this reading rests on a deliberate design choice, whose implications we discuss in Section~\ref{sec:limitations}.

\paragraph{Claude 3.5 Haiku: searches followed by abstentions.}
Haiku's indicator reports a search on 64\% of \OK{} items, yet it abstained on 64\% of those (79 of 124) and was correct on only 30\% of them. This is unlike the other models, whose searches nearly always yielded an answer. In 31 of these 79 abstentions the stored reason refers to a ``future'' or ``fictional'' date, which suggests that the search results did not reach the final answer, for example because of the 100-token output cap, the two-stage extraction (Section~\ref{sec:setup}), or because the indicator is a false positive. We therefore treat Haiku's NAS ($-0.15$) and KIS (0.42) as reflecting our harness and the vendor tool as a whole, and not a clean measure of the model's search decision. Haiku also searched on 11\% of \IK{} questions.

\paragraph{Gemini 2.5 Flash: search indicator is uninformative.}
In our harness the Gemini ``searched'' flag is set whenever the response has at least one candidate, which is always the case. It is therefore 1 for all 400 with-search runs and gives no information about whether Gemini searched. The resulting KIS of $-1$ must not be read as evidence that Gemini ``searches regardless of question type.'' Gemini's with-search NAS and KIS are therefore reported for completeness but \emph{excluded from all comparisons}. Verifying Gemini's behaviour requires re-running it with grounding metadata (the list of issued search queries), which we leave to future work.

\subsection{Search triggering and temporal cues}

A post-hoc check shows that search decisions are partly driven by surface cues. About 58\% of \OK{} stems contain an explicit temporal cue (a year such as 2025, ``November'', ``recent'', and similar; Table~\ref{tab:dataset_stats}), against 1 of 200 \IK{} stems. Table~\ref{tab:cues} splits the with-search \OK{} results by this cue. GPT-5 mini and Grok 4 Fast search on 98--99\% of cued stems but on only 77--82\% of uncued ones, and Mistral Small's rare searches are concentrated on cued stems (34\% vs.\ 2\%). Thus part of what NAS and KIS reward is sensitivity to the wording of the question, which the \IK/\OK{} split confounds with knowledge status. The scores should be read as measuring \emph{search triggering}, a behavioural proxy for metacognition, and not as direct evidence about models' internal self-knowledge. Restricting NAS to the uncued \OK{} items (83 per model, 80 for Haiku) gives lower and more variable values than on cued ones for three of the four models; Haiku, whose indicator is unverified, is the exception: GPT-5 mini 0.58 [0.43, 0.71] vs.\ 0.83 [0.75, 0.90], Grok 4 Fast 0.57 [0.42, 0.71] vs.\ 0.85 [0.78, 0.91], Mistral Small $-0.81$ [$-0.90$, $-0.71$] vs.\ $-0.28$ [$-0.45$, $-0.09$], and Claude 3.5 Haiku $-0.01$ [$-0.18$, $0.15$] vs.\ $-0.25$ [$-0.37$, $-0.11$]. The sign of NAS is unchanged for every model, but GPT-5 mini and Grok are clearly less selective than their headline NAS suggests when the question gives no temporal hint. The cue regular expression is crude and was defined post hoc.

\begin{table}[t]
\centering
\small
\begin{tabular}{lcc}
\toprule
Model & search rate, uncued \OK{} stems & search rate, cued \OK{} stems \\
\midrule
GPT-5 mini & 0.82 ($n$=83) & 0.98 ($n$=117) \\
Grok 4 Fast & 0.77 ($n$=83) & 0.99 ($n$=117) \\
Mistral Small & 0.02 ($n$=83) & 0.34 ($n$=108) \\
Claude 3.5 Haiku$^\ddagger$ & 0.66 ($n$=80) & 0.62 ($n$=114) \\
\bottomrule
\end{tabular}
\caption{Search rate on \OK{} questions with search, by presence of an explicit temporal cue in the stem. Gemini is omitted (constant indicator). $^\ddagger$Indicator unverified.}
\label{tab:cues}
\end{table}

\subsection{NAS and KIS}

\begin{figure}[t]
\centering
\begin{tikzpicture}
\begin{axis}[
  width=0.72\linewidth, height=7.2cm,
  xlabel={Net Awareness Score (NAS)}, ylabel={Knowledge Integrity Score (KIS)},
  xmin=-1, xmax=1, ymin=-1.1, ymax=1.1,
  grid=both, grid style={gray!20},
  tick label style={font=\footnotesize}, label style={font=\small},
  clip=false
]
\addplot[dashed, gray] coordinates {(0,-1.1) (0,1.1)};
\addplot[dashed, gray] coordinates {(-1,0) (1,0)};
\node[gray, font=\scriptsize, align=center] at (axis cs:0.5,0.18) {ideal};
\node[gray, font=\scriptsize, align=center] at (axis cs:-0.5,0.18) {under-searcher};
\node[gray, font=\scriptsize, align=center] at (axis cs:0.5,-0.18) {over-searcher};
\node[gray, font=\scriptsize, align=center] at (axis cs:-0.5,-0.18) {confused};

\addplot[only marks, mark=*, mark size=3pt, blue!70!black, mark options={solid,fill=blue!70!black}, error bars/.cd, x dir=both, x explicit, y dir=both, y explicit, error bar style={solid,blue!70!black}]
  coordinates {(0.72,0.79) +- (0.075,0.06)};
\node[anchor=south east, font=\scriptsize] at (axis cs:0.66,0.86) {GPT-5 mini};
\addplot[only marks, mark=triangle*, mark size=3.5pt, red!70!black, mark options={solid,fill=red!70!black}, error bars/.cd, x dir=both, x explicit, y dir=both, y explicit, error bar style={solid,red!70!black}]
  coordinates {(0.73,0.76) +- (0.07,0.065)};
\node[anchor=north west, font=\scriptsize] at (axis cs:0.55,0.64) {Grok 4 Fast};
\addplot[only marks, mark=diamond*, mark size=3.5pt, green!50!black, mark options={solid,fill=green!50!black}, error bars/.cd, x dir=both, x explicit, y dir=both, y explicit, error bar style={solid,green!50!black}]
  coordinates {(-0.51,0.80) +- (0.115,0.055)};
\node[anchor=south, font=\scriptsize] at (axis cs:-0.51,0.88) {Mistral Small};
\addplot[only marks, mark=square*, mark size=3pt, orange!80!black, mark options={solid,fill=orange!80!black}, error bars/.cd, x dir=both, x explicit, y dir=both, y explicit, error bar style={solid,orange!80!black}]
  coordinates {(-0.15,0.42) +- (0.10,0.095)};
\node[anchor=west, font=\scriptsize] at (axis cs:-0.03,0.42) {Claude 3.5 Haiku};
\addplot[only marks, mark=o, mark size=3.5pt, gray, mark options={solid}, error bars/.cd, x dir=both, x explicit, error bar style={solid,gray}]
  coordinates {(0.59,-1.0) +- (0.07,0)};
\node[anchor=south, font=\scriptsize, gray, align=center] at (axis cs:0.59,-0.93) {Gemini 2.5 Flash\\(indicator unreliable)};
\end{axis}
\end{tikzpicture}
\caption{NAS vs.\ KIS in the with-search condition. Bars are 95\% bootstrap intervals over questions (horizontal and vertical half-widths). Gemini's hollow marker is not interpretable because its search flag is constant. Quadrant names are informal.}
\label{fig:quadrant}
\end{figure}
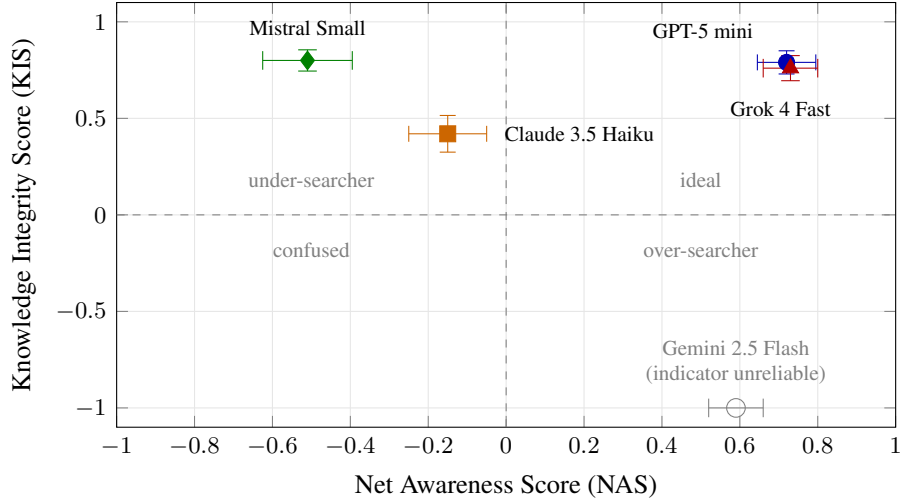

Table~\ref{tab:scores} and Figure~\ref{fig:quadrant} give the scores. In the with-search condition, GPT-5 mini and Grok 4 Fast are in the ``ideal'' region (NAS 0.72/0.73, KIS 0.79/0.76) and are statistically indistinguishable from each other. Mistral Small lies in the under-searcher region (NAS $-0.51$, KIS 0.80): it is restrained on \IK{} items but does not search on \OK{} items. Claude 3.5 Haiku has negative NAS ($-0.15$) and intermediate KIS (0.42). Gemini 2.5 Flash is not interpretable (see above). With only five models, we draw no conclusion about the relationship between model size, vendor or release date and search-triggering behaviour; the models differ simultaneously in architecture, training and search tooling.

\begin{table}[t]
\centering
\small
\begin{tabular}{lcccc}
\toprule
 & \multicolumn{2}{c}{Full definition} & \multicolumn{2}{c}{No reasoning judge} \\
\cmidrule(lr){2-3}\cmidrule(lr){4-5}
Model & NAS [95\% CI] & KIS [95\% CI] & NAS & KIS \\
\midrule
\multicolumn{5}{l}{\emph{With search (primary)}} \\
GPT-5 mini        & 0.72 [0.65, 0.80]  & 0.79 [0.73, 0.85] & 0.74 & 0.98 \\
Grok 4 Fast       & 0.73 [0.66, 0.80]  & 0.76 [0.69, 0.82] & 0.73 & 0.97 \\
Mistral Small     & $-$0.51 [$-$0.62, $-$0.39] & 0.80 [0.74, 0.85] & $-$0.38 & 0.99 \\
Claude 3.5 Haiku  & $-$0.15 [$-$0.25, $-$0.05] & 0.42 [0.33, 0.52] & $-$0.10 & 0.76 \\
Gemini 2.5 Flash$^\dagger$ & 0.59 [0.52, 0.66]  & $-$1.00 & 0.59 & $-$1.00 \\
\midrule
\multicolumn{5}{l}{\emph{Without search (secondary)}} \\
GPT-5 mini        & 0.62 [0.52, 0.72] & 0.65 [0.54, 0.75] & 0.72 & 0.98 \\
Grok 4 Fast       & 0.80 [0.71, 0.86] & 0.50 [0.37, 0.62] & 0.87 & 0.99 \\
Mistral Small     & 0.41 [0.31, 0.52] & 0.53 [0.41, 0.64] & 0.65 & 0.98 \\
Claude 3.5 Haiku  & 0.79 [0.71, 0.86] & 0.58 [0.46, 0.69] & 0.90 & 0.97 \\
Gemini 2.5 Flash  & 0.61 [0.52, 0.70] & 0.52 [0.40, 0.64] & 0.81 & 0.99 \\
\bottomrule
\end{tabular}
\caption{NAS (\OK{} items) and KIS (\IK{} items). The last two columns drop the reasoning-alignment criterion (every correct answer is treated as aligned). Intervals bootstrap questions only. $^\dagger$Search indicator unreliable; not interpretable.}
\label{tab:scores}
\end{table}

\subsection{Sensitivity to the reasoning-alignment judge}

The last two columns of Table~\ref{tab:scores} show how much the scores depend on the LLM-judged reasoning criterion. Without it, KIS rises to 0.97--0.99 for every model with search except Claude 3.5 Haiku (0.76) and Gemini. With it, KIS is 0.76--0.80 for GPT-5 mini, Grok 4 Fast and Mistral Small, and 0.50--0.65 in the no-search condition. The judge rates only 75--83\% of correct \IK{} answers as aligned for most models (62\% for Claude 3.5 Haiku with search and 58\% for Gemini with search), so most of the ``knowledge integrity'' gap to 1 is attributable to the judge, not to wrong answers or unnecessary searches. We have not checked the judge against human annotation, and short, 100-token reasons may be judged harshly. KIS should therefore be read as a conservative composite and the judge-free version as a measure of restraint and accuracy only. In the with-search condition the ordering by NAS is essentially unchanged by the criterion, except that Mistral's NAS is less negative ($-0.38$), and the qualitative picture of Table~\ref{tab:breakdown} does not depend on it. Without search, NAS shifts by 0.07--0.24 and the ordering changes (e.g., GPT-5 mini and Gemini swap places), so no-search NAS depends materially on the judge.

\subsection{Without search}

Without tools, the appropriate response to an \OK{} question is abstention. Abstention rates on \OK{} items were 77\% (Claude 3.5 Haiku), 70\% (Grok 4 Fast), 59\% (Gemini 2.5 Flash), 50\% (GPT-5 mini) and 43\% (Mistral Small), against $\le$0.5\% on \IK{} items. All models thus abstain far more on \OK{} than on \IK{} items, which indicates some ability to tell the two apart, but for GPT-5 mini and Mistral Small about half of \OK{} questions were answered anyway, usually correctly (Table~\ref{tab:acc}). Because the correct-with-aligned-reasoning case counts as aware, no-search NAS (0.41--0.80) mostly reflects abstention rate and is only moderately informative; the no-search condition is best seen as a control for the with-search results.

\section{Limitations and Threats to Validity}
\label{sec:limitations}

This is a preliminary study, and the following issues limit what can be concluded.

\paragraph{Measurement of search.}
Search indicators are vendor-specific and derived from API output structure. We verified by code inspection that they are plausible for four of the models, but Gemini's is not (Section~\ref{sec:results}). The with-search system prompt discourages search except for current or obscure information, and the results may differ under other prompts or vendor defaults. Vendor search stacks also differ in retrieval quality, which confounds model comparisons.

\paragraph{Score design.}
(i) With search, abstention without searching counts as unaware, which penalises models like Mistral that detect the gap but do not act. A more nuanced score would separate \emph{noticing} from \emph{acting}. (ii) KIS is asymmetric across conditions (Section~\ref{sec:metrics}). (iii) The sign of NAS and KIS and the quadrant boundaries are conventions. (iv) The reasoning-alignment judge is unvalidated and has a large effect on KIS (Table~\ref{tab:scores}).

\paragraph{Item validity.}
\IK{} items are near ceiling (98.5--99.5\% accuracy), so they discriminate models mainly by their search behaviour, not by knowledge. Without search, models that answered \OK{} items were right 69--79\% of the time, well above chance, so some \OK{} items are partly answerable by priors or by option elimination; for example, a question about which precious metal's price fell by a given amount in rupees per 10 grams contains enough cues for an informed guess (Appendix~\ref{sec:question_examples}). Some \OK{} stems refer to ``the article'' and cannot be answered even with search, and about 12\% of \OK{} items stay unanswered for the strongest models even after search. Theme extraction also causes \emph{domain drift}: an \IK{} question labelled ``entertainment'' may be about arms-trade treaties, so per-domain results (which rest on 50 items per cell) should be read cautiously, and we do not report them. Questions were generated and validated by single LLMs (GPT-4o-mini, gpt-5-mini), which may impose stylistic biases, and no human validation of items was performed. All articles are from two consecutive days, so the benchmark samples narrow temporal and topical coverage.

\paragraph{Harness and prompt inconsistencies.}
The prompts delivered to models differ: GPT-5 mini did not receive the shared system prompt or the search-discouraging sentence, and Grok's with-search system prompt contained an irrelevant sentence (Section~\ref{sec:setup}). Three models use a two-stage answer extraction, and Claude's search indicator and first-stage text selection were not validated against raw API responses; Haiku's high rate of abstention after searching is therefore uninterpretable as model behaviour. The scoring notebook in the repository does not drop failed runs (it tags them $-1$ and combines tags with bitwise operators), so the paper's tables were recomputed with failures excluded using a separate script that we release with the repository; because only 17 runs failed, the difference is small. Temporal cues in the stems confound the \IK/\OK{} contrast (Section~\ref{sec:results}).

\paragraph{Answer-key errors.}
Some \OK{} keys are wrong or time-dependent. For example, the factoid of one \OK{} item about the second season of \emph{Landman} states a premiere on ``Sunday, November 18, 2025'', but 18 November 2025 was a Tuesday, and a question such as ``when did the stock reach its highest price in the past year'' (Appendix~\ref{sec:question_examples}) is only correct on the publication date. We did not audit the full set, so some of the abstentions after search (about 12\% of \OK{} items for the strongest models) are probably due to such items rather than to model failure.

\paragraph{Contamination.}
Although the articles post-date the vendors' stated cut-offs, facts may still be partially predictable from earlier coverage, and we cannot verify the training data of any model.

\paragraph{Scope and statistics.}
Five models, one run per item and condition, and 200 items per type yield wide intervals (about $\pm$0.07 for with-search NAS and KIS, up to $\pm$0.12 for no-search KIS) that do not account for run-to-run stochasticity. The MCQ format also constrains how uncertainty can be expressed, and the evaluated ``agent'' makes a single search-or-not decision, whereas real agents plan over multiple steps. The model choice mixes a 2024 model (Claude 3.5 Haiku) with late-2025 models, so differences cannot be attributed to a single factor.

\section{Conclusion and Future Work}

We introduced an inside/outside-knowledge benchmark and two action-level scores for evaluating whether search-enabled LLMs act appropriately on the boundary of their knowledge. In a small study of five commercial products, GPT-5 mini and Grok 4 Fast searched selectively; Mistral Small mostly failed to search on \OK{} questions even when it abstained; Claude 3.5 Haiku's searches were frequently followed by abstentions, probably due to our harness; and Gemini 2.5 Flash's behaviour could not be measured because of a defective search indicator. A sensitivity analysis showed that with-search NAS is robust to the judge-based criterion whereas KIS is not, and a post-hoc analysis showed that temporal cues in the questions partly drive search. These results are therefore indicative of search-triggering behaviour in specific product configurations, and not definitive evidence about the metacognition of the underlying models.

The most valuable next steps are (i) re-running Gemini with grounding metadata to recover its search decisions, (ii) validating the reasoning-alignment judge against human labels or replacing it with a judge-free KIS, (iii) separating \emph{noticing} a gap (abstention or verbalised uncertainty) from \emph{acting} on it (search) in the scoring, (iv) pruning \OK{} items that are answerable without the article, (v) repeating runs and varying the search instruction to estimate run-to-run and prompt sensitivity, and (vi) extending to open-ended questions, elicited confidence and multi-step agent workflows.

\bibliographystyle{plainnat}
\bibliography{refs}

\appendix
\section{Example Questions}
\label{sec:question_examples}

Each item stores additional metadata (source article, factoid, paraphrase, validation scores); we show the main fields. Full records are in the repository.

\subsection{Example \IK{} question}
\begin{verbatim}
question: What is the primary international agreement that regulates
          the trade of conventional arms to ensure it does not
          contribute to human rights violations and armed conflict?
choices : 1 The Landmines Convention (Ottawa Treaty)
          2 The Comprehensive Nuclear-Test-Ban Treaty (CTBT)
          3 The Non-Proliferation Treaty (NPT)
          4 The Arms Trade Treaty (ATT)          <- correct
reason  : The Arms Trade Treaty aims to establish common standards for
          the international trade of conventional weapons and to
          prevent their diversion to the illicit market or use in
          human rights abuses.
domain  : entertainment   (source article: US politics live blog;
                           the theme, not the article, drives the question)
\end{verbatim}
This item illustrates \emph{domain drift}: the question is generic and unrelated to the domain label of its source article.

\subsection{Example \OK{} questions}
\begin{verbatim}
question: What date did the stock price of Monster Beverage
          Corporation reach its highest point in the past year?
choices : 1 July 30, 2025   2 March 22, 2025
          3 January 15, 2025   4 November 18, 2025   <- correct
factoid : On November 18, 2025, Monster Beverage Corporation's stock
          price reached a new 52-week high.
domain  : business     relevance: 10     correctness: [0,0,0,10]
\end{verbatim}
\begin{verbatim}
question: What precious metal saw a price decrease of Rs 3,900,
          reaching Rs 1,25,800 for every 10 grams in the National
          Capital on November 18, 2025?
choices : 1 Platinum   2 Gold   3 Silver   4 Palladium
correct : 2 (Gold)     domain: technology
\end{verbatim}
Both items concern facts from the source articles, but the stems carry cues (the most recent date among the distractors; a price level typical of gold) that make informed guessing possible, which is the item-validity issue discussed in Section~\ref{sec:limitations}. The second is also labelled ``technology'' though it concerns commodity prices.

\section{Dataset Generation Prompts}
\label{sec:dataset_generation_prompt}

Through trial and error, it was found that for the structured generation tasks, few-shot examples noticeably limit the variety of question styles and topics as well as bias the model towards more verbose outputs that later fail the format validation process. Thus few-shot examples were not used in prompts for the generation of the final submitted datasets, except for the theme extraction.

\subsection{Factoid Generation Prompt}

\begin{verbatim}
    You are an assistant that doesn't make mistakes. Given the 
    following news excerpt and publication, extract one fact from 
    this article. This factual statement should contain the exactly 
    correct information with no mistakes. The meaning should be clear 
    and unambiguous without any addition context. The factual 
    statement should contain enough information so that a non-
    specialist reader can understand the statement. Any reference to 
    potentially unknown entity must be briefly explained. If the 
    statement is time-sensitive, include day, month, or year required 
    to understand the statement. Any time-sensitive statement (such 
    as "last Tuesday" or "5 years from now") must be given 
    appropriate and accurate time of reference.

    The news excerpt: {title + description + content of news article}

\end{verbatim}

\subsection{Question Generation Pipeline Prompts}

\subsubsection{Paraphrase Prompt}

\begin{verbatim}

    Paraphrase the given context: {{ {factoid} }}

\end{verbatim}

\subsubsection{Keyword Extraction Prompt (for OK)}

\begin{verbatim}

    Extract one keyword which contain important factual information 
    from the paraphrased context {{ {paraphrase1} }}. Pick a keyword 
    which represents a specific information, rather than generic 
    terms.

\end{verbatim}

\subsubsection{Theme Extraction Prompt (for IK)}

\begin{verbatim}
    Determine a theme of the paraphrased context {{ {paraphrase1} }}. 
    Output one to two words high-level theme of the provided 
    text. For example : 'baseball' or 'business merger'
\end{verbatim}

\subsubsection{Question Generation Prompt (for OK)}

\begin{verbatim}
    Generate a question based on the paraphrased context 
    {paraphrase1} and the correct answer {keyword}. Include only the 
    question and nothing else in your response.
\end{verbatim}

\subsubsection{Question Generation Prompt (IK)}

\begin{verbatim}
    Generate a quiz question and a short, correct answer based on the 
    theme {keyword}. Include only the question and nothing else in 
    one line, and only the answer and nothing else in the next line. 
    No colons, no bulletpoints, no periods. No time-sensitive 
    information may be used in the question. Briefly explain the 
    reasoning for the answer on the next line.
\end{verbatim}

\subsubsection{Distractor Generation Prompt}

\begin{verbatim}
    Create three plausible distractors for the question 
    {{ {question} }} and the correct answer {{ {keyword (for OK)} / 
    {correct_answer (for IK)} }}. 
    Each distractor should be on separate lines. No numbering, no 
    bulletpoints, no periods. The distractors should be unambiguously 
    false given the context {{ {paraphrase1} }}. The three 
    distractors should be distinct from each other and from the 
    correct answer { and all answer choices should have similar 
    lengths (for IK) }.
\end{verbatim}

\subsection{Validation Prompts}

\subsubsection{Format Revision Prompt}

\begin{verbatim}
    You are a professional assistant who never makes mistakes. Your 
    job is to revise quiz questions so that they are formatted 
    correctly and consistently. You will revise ONLY:
    1. The question text
    2. The correct answer choice

    You will be given:
    - The question
    - The correct answer choice
    - The other correct choices (to ensure consistent capitalization, 
    punctuation, and other formatting quirks between the correct 
    answer choice and distractors)

    Do not modify other choices.
    Do not modify the meaning.
    Remove extraneous prefixes (e.g., "Question:", "Answer:", 
    "Correct:", etc.).
    Ensure consistent capitalization and punctuation.
    Remove periods for answer choices.
    Do NOT add new content.
    Do NOT rewrite incorrect choices.
    Do NOT change the correct choice number.

    Return ONLY the following JSON object:

    {{
    "question": "...",
    "correct_answer": "..."
    }}

    Here is the input:

    Question: {question}

    Choices:
    1. {c1}
    2. {c2}
    3. {c3}
    4. {c4}

    Correct answer choice number: {correct_index}
    Correct answer text: {correct_answer}
\end{verbatim}

\subsubsection{Format Validation Prompt}

\begin{verbatim}
    You are a professional assistant who never makes mistakes. Your 
    job is to evaluate whether a quiz question is formatted correctly 
    and consistently.

    A correctly formatted question contains a question stem and 4 
    answer choices.
    A consistently formatted question has answer choices with the 
    same style of capitalization, punctuation, etc., such that no 
    answer choice stands out just by virtue of its stylistic features.
        
    Given a question, determine if the formatting is correct AND 
    consistent.
    
    Return ONLY "Yes" or "No"
    
    Here is the input:
    
    Question: {question}
    
    Choices:
    1. {c1}
    2. {c2}
    3. {c3}
    4. {c4}
    
    Correct answer choice number: {correct_index}
    Correct answer text: {correct_answer}
    
\end{verbatim}

\subsubsection{Grammar Validation Prompt}

\begin{verbatim}
    Consider the following question and answer choices:

    Question: {question}

    Choices:
    1. {c1}
    2. {c2}
    3. {c3}
    4. {c4}

    Is there any grammatical error in the question? Answer ONLY 
    'Yes' or 'No'.
\end{verbatim}

\subsubsection{Relevance Validation Prompt (for OK only)}

\begin{verbatim}
    Considering the following article and the question based on 
    this article:

    {factoid}

    Question: {question}

    Please grade the relevance of this question relative to the text 
    with a score from 0 to 10. 
    A question is considered relevant if it pertains to the core 
    themes and concepts discussed in the text, 
    engages with the important ideas and content presented, and does 
    not focus on memorizing overly-specific details 
    such as precise wording. 

    Answer ONLY with the number corresponding to the score.
\end{verbatim}

\subsubsection{Answer Scoring Prompt (called once for each possible answer choice)}

\begin{verbatim}
    You are a professional assistant who never makes mistakes. You 
    will be given a question and a potential answer. Your job is to 
    rate how correct the answer is, relative to the question and any 
    other information given in the input.

    Question: {question}

    Answer choice: {answer}

    {factoid_section}

    Rate how correct this answer is for the question on a scale from 
    0 to 10.
    - 0 = completely incorrect
    - 10 = perfectly correct

    Return ONLY a number from 0 to 10.
\end{verbatim}

\section{Evaluation Prompts}

\subsection{Answering prompt}
\label{sec:mcq_prompt}

These are the prompts built by the evaluation harness. As described in Section~\ref{sec:setup}, GPT-5 mini did not receive the system prompt in either condition (only the user prompt and a JSON-formatting sentence), and Grok 4 Fast's with-search system prompt had an irrelevant ``extract the invoice data into JSON'' sentence appended.

System prompt (all conditions):
\begin{verbatim}
Answer following multiple choice question based on your knowledge.
Provide the answer choice (1, 2, 3, 4 or 5) along with a very short
reason for your choice.
\end{verbatim}
Appended to the system prompt in the with-search condition:
\begin{verbatim}
**Only use the web search tool if the user asks for current, real-time
information or for obscure facts you do not know.**
\end{verbatim}
User prompt:
\begin{verbatim}
Question: {question}
Options:
1: {choice-1}
2: {choice-2}
3: {choice-3}
4: {choice-4}
5: Not enough information to answer
\end{verbatim}
Output schema: JSON with an integer field \texttt{choice} in 1--5 and a string field \texttt{reason}.

\subsection{Reasoning-alignment judge prompt} \label{sec:reasoning_prompt}

\begin{verbatim}
You are an expert evaluator of reasoning alignment. 
Your task is to assess how closely a model-generated reason matches a ground-truth reason.

You must evaluate **semantic equivalence**, not surface similarity.

=====================
EVALUATION PROTOCOL (SAE)

1. NORMALIZATION
Break both reasons into their fundamental propositions or meaning units.

2. ALIGNMENT
For each ground-truth proposition:
- Determine whether the model reason expresses the same idea.
- Mark it as: "Fully Matched", "Partially Matched", or "Not Matched".

3. SCORING
Compute a semantic alignment score from 0 to 100:
- 100 = meanings fully equivalent  
- 70-99 = mostly same meaning with minor omissions  
- 40-69 = partial overlap, several missing or incorrect elements  
- 1-39 = mostly different  
- 0 = completely unrelated

4. ERROR ANALYSIS
List:
- Missing key ideas
- Incorrect additions
- Distortions or misunderstandings

=====================
GROUND TRUTH REASON:
{grd_reason}

MODEL-GENERATED REASON:
{gen_reason}
=====================
OUTPUT FORMAT (JSON ONLY)
{
"score": <0-100> (integer) **Do not verbalize the score. Ex: fifty -> 50**,
"alignment": 0 or 1 (1 = aligned if score >= 70, else 0),
"summary": "<short evaluator summary>"
}
\end{verbatim}

\section{Additional Dataset Statistics}

\begin{figure}[h]
    \centering
    \begin{subfigure}{0.23\linewidth}
        \centering
        \includegraphics[width=\linewidth]{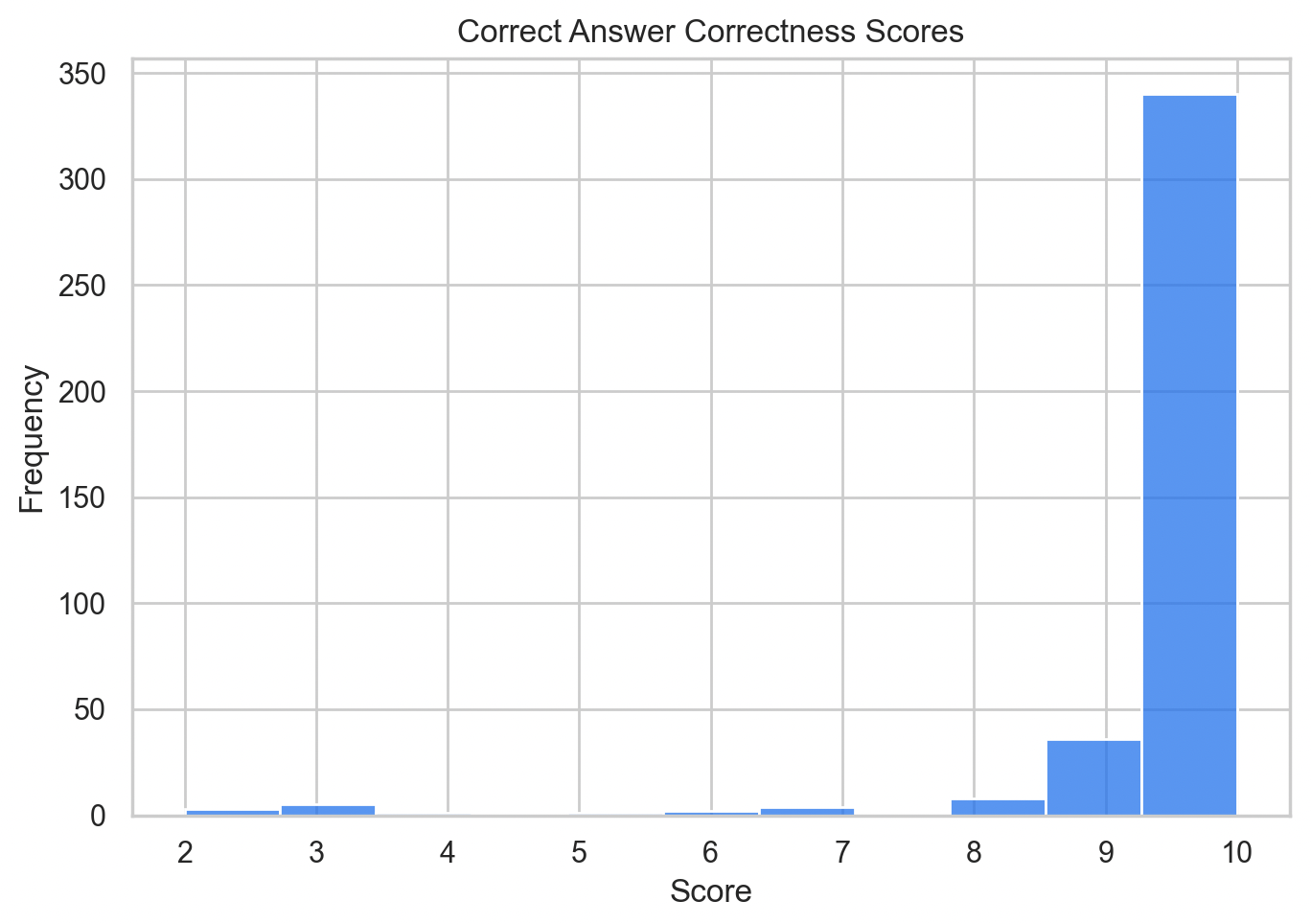}
        \caption{Correctness of keyed options}
    \end{subfigure}\hfill
    \begin{subfigure}{0.23\linewidth}
        \centering
        \includegraphics[width=\linewidth]{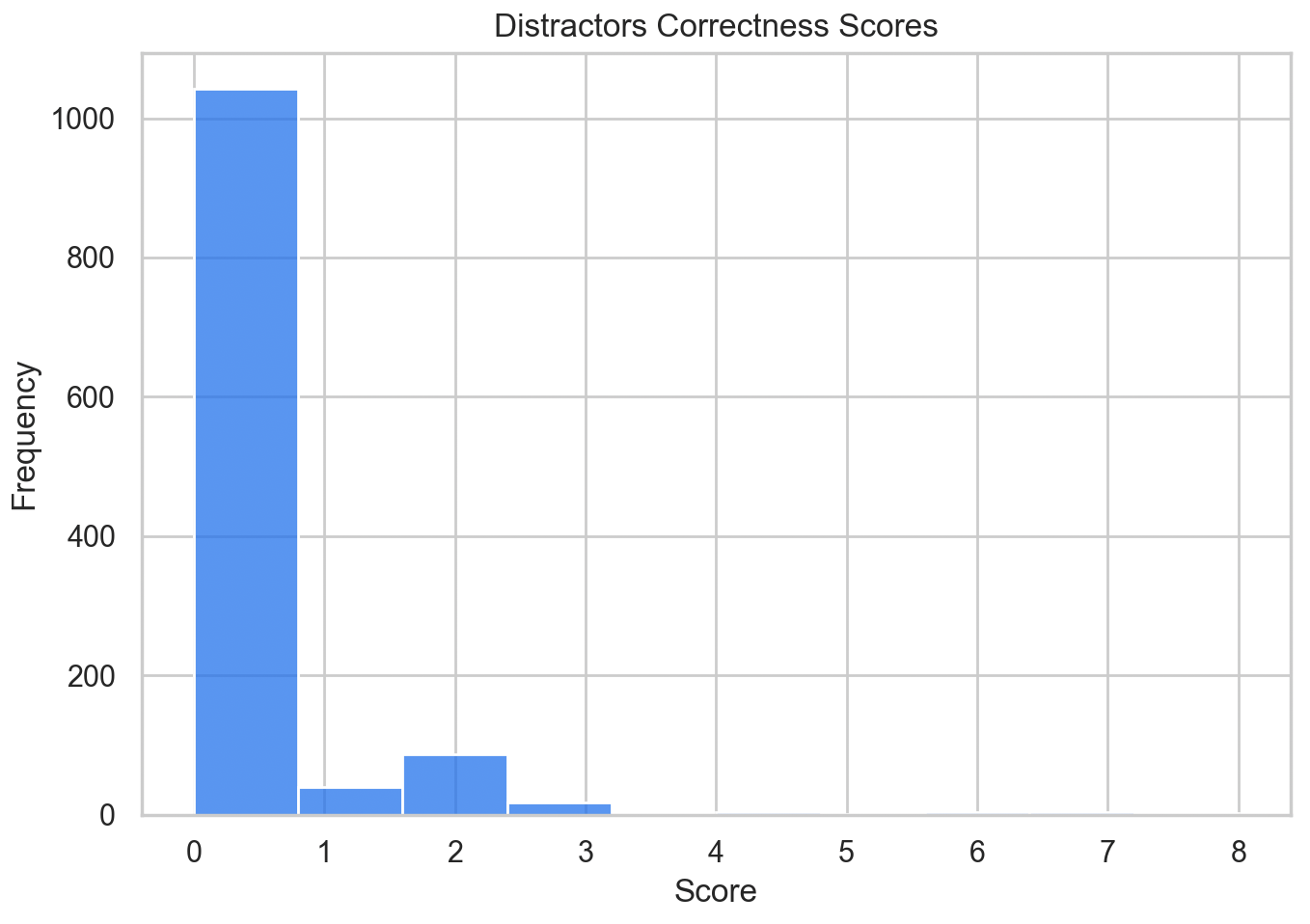}
        \caption{Correctness of distractors}
    \end{subfigure}\hfill
    \begin{subfigure}{0.23\linewidth}
        \centering
        \includegraphics[width=\linewidth]{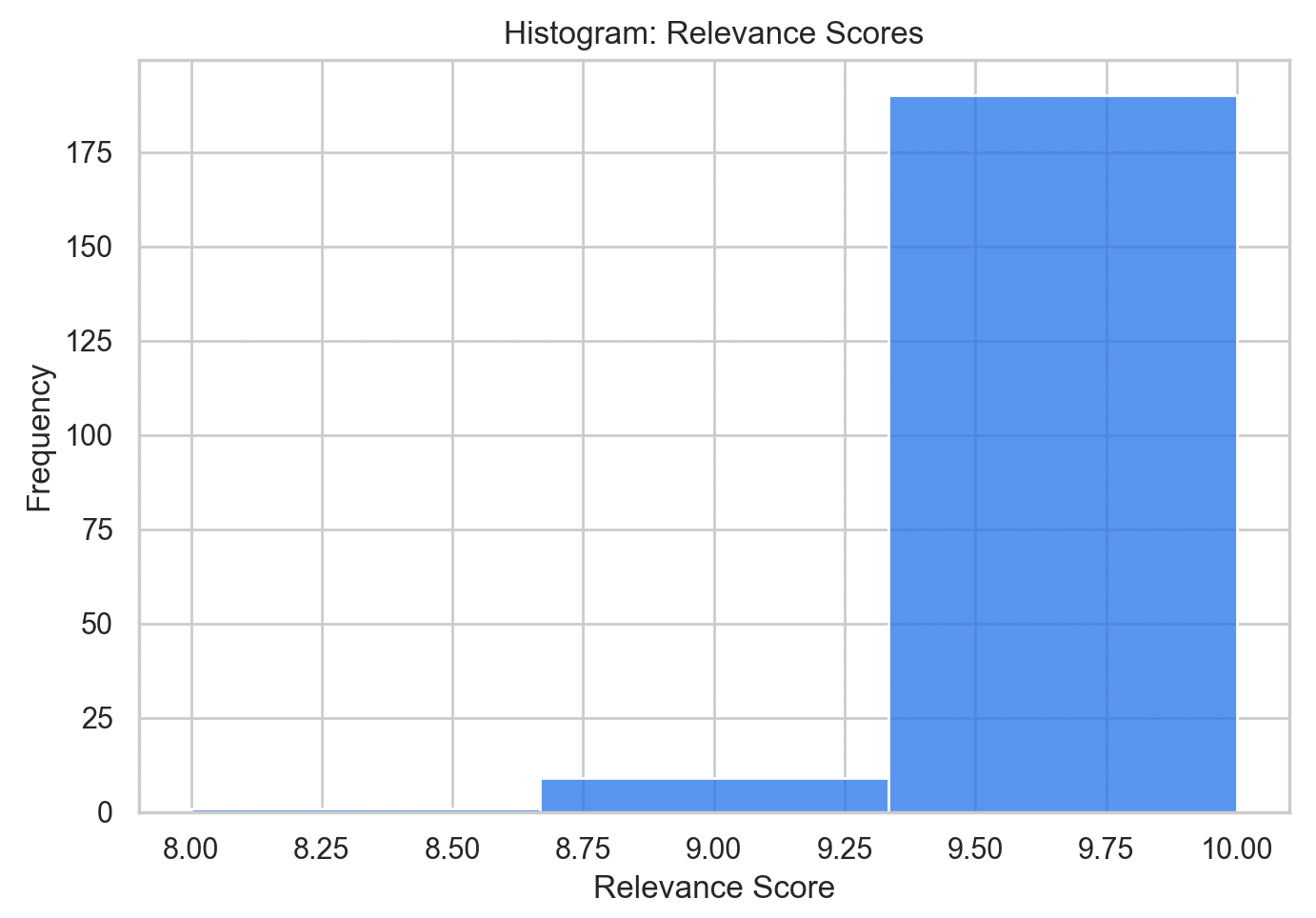}
        \caption{Relevance, \OK{} items}
    \end{subfigure}\hfill
    \begin{subfigure}{0.23\linewidth}
        \centering
        \includegraphics[width=\linewidth]{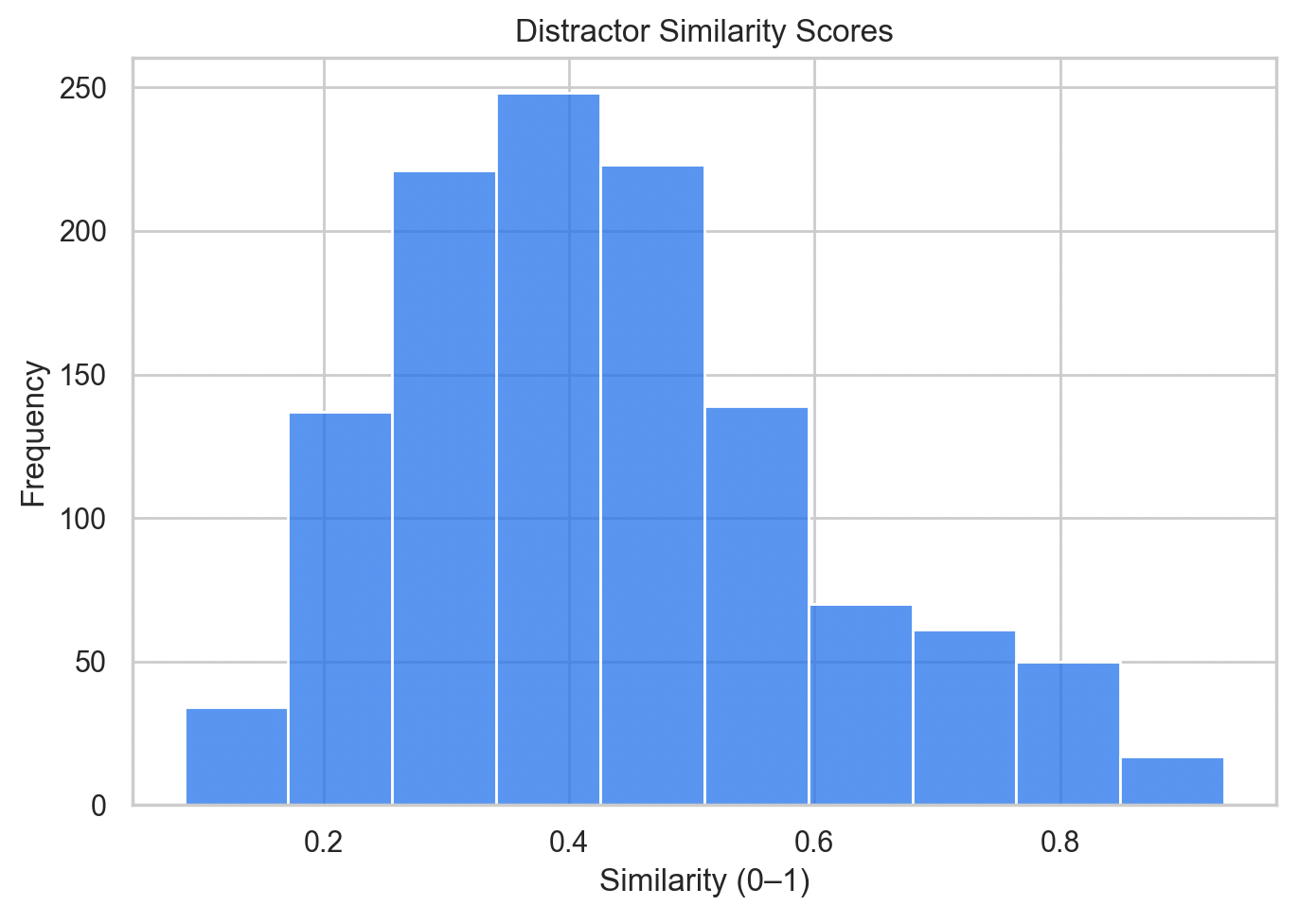}
        \caption{Distractor--answer cosine similarity}
    \end{subfigure}
    \caption{Distributions of validation scores for the final dataset.}
    \label{fig:combined}
\end{figure}

\end{document}